\documentclass[conference]{IEEEtran}
\IEEEoverridecommandlockouts
\usepackage{comment}
\usepackage{amsmath,amsfonts}
\usepackage{array}
\usepackage[caption=false,font=normalsize,labelfont=sf,textfont=sf]{subfig}
\usepackage{textcomp}
\usepackage{stfloats}
\usepackage{url}
\usepackage{verbatim}
\usepackage{graphicx}
\usepackage{array}
\usepackage{caption} 
\usepackage{xcolor}
\usepackage{multirow}
\usepackage[ruled, lined, linesnumbered, longend]{algorithm2e}
\usepackage{soul} 

\def\BibTeX{{\rm B\kern-.05em{\sc i\kern-.025em b}\kern-.08em
    T\kern-.1667em\lower.7ex\hbox{E}\kern-.125emX}}
    
\newcommand{\fixme}[1]{{\color{red}\em\bf{[FIXME: #1]}}}

\SetKwInput{KwOutput}{Output}   
\SetKwInput{KwInput}{Input}

\usepackage{fancyhdr}
\usepackage{kantlipsum}
\fancypagestyle{plain}{
\fancyhf{}
\fancyhead[C]{Conference on \LaTeX} 

}
\usepackage{eso-pic}
\begin{document}

\title{DEAP-FAKED: Knowledge Graph based Approach for Fake News Detection}

\author{\IEEEauthorblockN{Mohit Mayank}
\IEEEauthorblockA{
Outplay, India \\
mohit.m@outplayhq.com}
\and
\IEEEauthorblockN{Shakshi Sharma}
\IEEEauthorblockA{\textit{Institute of Computer Science} \\
University of Tartu, Estonia\\
shakshi.sharma@ut.ee}
\and
\IEEEauthorblockN{Rajesh Sharma}
\IEEEauthorblockA{\textit{Institute of Computer Science} \\
University of Tartu, Estonia\\
rajesh.sharma@ut.ee}
}

\maketitle
\begin{abstract}
    Fake News on social media platforms has attracted a lot of attention in recent times, primarily for events related to politics (2016 US Presidential elections), and healthcare (infodemic during COVID-19), to name a few. Various methods have been proposed for detecting Fake News. The approaches span from exploiting techniques related to network analysis, Natural Language Processing (NLP), and the usage of Graph Neural Networks (GNNs). In this work, we propose DEAP-FAKED, a knowleDgE grAPh FAKe nEws Detection framework for identifying Fake News. Our approach combines natural language processing (NLP) and tensor decomposition model to encode news content and embed Knowledge Graph (KG) entities, respectively. A variety of these encodings provides a complementary advantage to our detector. We evaluate our framework using two publicly available datasets containing articles from domains such as politics, business, technology, and healthcare. As part of dataset pre-processing, we also remove the bias, such as the source of the articles, which could impact the performance of the models. DEAP-FAKED obtains an F1-score of 88\% and 78\% for the two datasets, which is an improvement of $\sim$21\%, and $\sim$3\%, respectively, which shows the effectiveness of the approach.
\end{abstract}

\footnote{This is a preprint version of the accepted paper in IEEE/ACM International Conference on Advances in Social Networks Analysis and Mining (ASONAM) 2022.}
\section{Introduction}\label{sec:intro}
\vspace{-2pt}

Online social media platforms such as Twitter have become de facto news sources for the public in general \cite{boczkowski2017incidental}. However, not every piece of information escalated on these platforms is genuine. These platforms often cater to the spread of misinformation (such as Fake News, hoaxes, and rumors) \cite{flintham2018falling}, which is intended to deceive the readers for personal advantage deliberately. 
It is evident by the impact of misinformation on various events that Fake News may have a global impact; for instance, a widely circulated piece of Fake News about the 2016 US presidential election \cite{bovet2019influence} worried the world. Furthermore, the covid19 pandemic \cite{orso2020infodemic} has become a significant source of misinformation these days. It is imperative that misinformation can be found in practically every sphere, including health \cite{waszak2018spread}, politics \cite{guess2018selective}, and finance \cite{polak2012misinformation}.
This clearly reflects the utmost concern about proposing mechanisms for detecting misinformation.
\urldef{\urlA}\url{https://blog.google/products/search/introducing-knowledge-graph-things-not/}

Previously, researchers have looked into various aspects of misinformation, such as examining user profiles involved in rumors \cite{shu2019role} and determining the veracity of the rumor \cite{devi2020veracity}. Several techniques have been employed in this domain, for instance, NLP \cite{ahmed2017detection}, network-based approaches ranging from simple network analysis \cite{shu2019studying} to recent advancements of GNNs \cite{ShakshiIJCNN2021}, and multi-modal approaches as well \cite{wang2018eann}. Recently, heterogeneous graphs have been examined, such as exploiting social context information \cite{nguyen2020fang}.


In this work, we exploit a Knowledge Graph-based (KG) framework for detecting Fake News articles, a specific category of misinformation. KGs are the integrated graph-structured knowledge base that has been compiled from diverse sources\footnote{\urlA} consisting of multiple facts. Specifically, one fact is represented using \textit{head, relation, and tail} triplets $(h, r, t)$, where $h$ and $t$ are the nodes representing entities, and $r$ represents the relation between the two nodes (entities). 
In addition to storing the extracted knowledge, KGs are widely employed in machine learning for predicting tasks, as the usage of KGs has been shown in the literature to aid in the development of better models \cite{zhan2018loan}.


We have proposed DEAP-FAKED, a dual-part KG-based framework for Fake News detection. In the first part, we employ an NLP-based technique to encode the fake (true) news title. 
In our case, we only encode news titles, as we want to consider the minimal amount of information to perform predictions. 
The reason behind this approach is that by reducing the dependency on multiple attributes, this framework can be used for tasks such as early detection of Fake News wherein comments and other metadata are unavailable or difficult to extract.
For this purpose, we use biLSTM based neural networks, which perform well for small sized sequence data encoding. In this paper, we show that the standalone NLP technique for fake news detection can be enhanced if combined with entity-level interactions present in the KG. For this, we need to map entities present in the news title with the entities in the KG. In the second part, we first identify and extract named entities from the news text using Named-Entity Recognition (NER) and then map them to a KG using Named Entity Disambiguation (NED). After mapping, we use ComplEx embedding, a tensor decomposition model, to embed the KG entities into vector form. Finally, we combine the embeddings from the two parts to detect Fake News.

In comparison to previous works, our work is different in the following ways. First, we show that by utilizing only the titles of the news articles in our framework, we are able to achieve better results. Second, we have also handled the 
biasedness\footnote{https://hbr.org/2019/10/what-do-we-do-about-the-biases-in-ai}, which has largely been ignored in previous works. To evaluate our methodology, we used two publicly available datasets containing $\sim$15k articles in total. In comparison to the baseline methodologies we employed, we show that our framework produces better results consistently.  
Our framework is able to achieve an F1-score of 88\% and 78\% for the two datasets, which is an improvement of 21\% and 3\%, respectively.

The rest of the paper is organized as follows. Section \ref{sec:related} covers Related Work, Section \ref{sec:method} covers the Proposed Methodology, Section \ref{sec:experiment} covers Dataset Description and Experimental Setup, Section \ref{sec:Eval} covers Evaluation Results, and Section \ref{sec:concl} covers Conclusions and Future Work.


 %
\section{Related work}\label{sec:related}

This section first discusses literature with respect to misinformation covering both Fake News and rumors. Next, we discuss the role of KGs in misinformation detection.

Presently, misinformation has attracted a lot of attention from the research community, and it has been examined from a variety of angles, in particular, utilizing user profiles to identify users who are involved in rumors \cite{shu2019role}, determining the veracity of rumors on social media platforms \cite{devi2020veracity}, and fact-checked data to identify fake news on health focusing on COVID-19 vaccine \cite{sharma2022facov}. 
Researchers have used techniques such as NLP \cite{sharma2021misleading, butt2022goes}, network-based, ranging from simple network analysis \cite{shu2019studying} to exploiting current breakthroughs in GNNs such as utilizing Graph Convolutional Networks \cite{ShakshiIJCNN2021} and gated GNN \cite{zhang2020fakedetector}. Apart from utilizing text, authors also looked into a multi-modal approach \cite{dhawan2022game,jagtap2021misinformation}.

Recently, heterogeneous graphs are also being explored, which is a promising direction in misinformation detection, especially in Fake News.
In \cite{nguyen2020fang}, authors detect Fake News leveraging social context information into a heterogeneous graph. News articles and their metadata are being used to build heterogeneous information networks \cite{ren2020hgat}. Heterogeneous Graph Neural Networks such as Adversarial active learning \cite{ren2020adversarial} and Graph-aware co-attention networks \cite{lu2020gcan} have also been explored for detecting Fake News.
As part of the heterogeneous graphs, KGs-based approaches have also been investigated. For example, a Knowledge-driven Multi-modal Graph Convolutional Network (KMGCN) has been used in \cite{wang2020fake}. In \cite{tchechmedjiev2019claimskg}, by harvesting data from popular fact-checking websites and exploring additional information from DBpedia, the authors introduced a KG of fact-checked claims. 

Our work differs from other works in a multitude of ways. First, we use minimal information (such as only the title) for the classification task. The reason behind this approach is in some cases, such as in the early detection of fake news, comments and other metadata are unavailable or difficult to extract. Second, we're trying to identify and subsequently eliminate any biases that may exist in the datasets. We compare the model's performance with and without biases in the dataset, and we discover that removing biases improves the model's performance. The KG-related works are not comparable to ours since they either employ other KGs or use KGs in conjunction with GNN techniques. 

%
\section{Methodology}\label{sec:method}

Our proposed framework, DEAP-FAKED, consists of the following three components. 
\begin{enumerate}
    \item \textbf{News encoder:} this component performs the contextual encoding of the news title.
    \item \textbf{Entity encoder:} this component identifies the named entities present in the news title and encodes the individual entities using KG. 
    \item \textbf{Classification Layer:} this component consolidates the news encoder's and entity encoder’s representations to perform the final downstream Fake News classification learning.
\end{enumerate}


\subsection{News encoder}

The efficient representation of sequential data is a long-standing research problem in the NLP domain \cite{torfi2020natural}. The intention is to represent the text data, which is inherently sequential in nature, into a continuous vector representation. 
While conventional work has focused on the sequential representation of textual data, which is unidirectional in nature, recent work \cite{li2020survey} proposed a more efficient contextual representation, which is bidirectional. 
We tried a variety of stacking \cite{nkenyereye2021stacking}, unidirectional and bidirectional sequence encoders \cite{bahad2019fake}. Finally, we select a 2-layer stacked biLSTM as the main subcomponent of our news encoder. 
Each token is represented using $t_i$ where $1 \leq i \leq n$.
The size of the tokens ($n$) depends on the news title's length and is limited to a maximum of 256 tokens. 

\subsection{Entity encoder}

The recent trend in Fake News detection research leverage complementary information, apart from the news title, to further improve the detection performance. 
We leverage inherent information present within the news text - \textit{Named Entities}. For example, a news title with the text - “\textit{US Officials See No Link Between Trump and Russia}" contains two entities - “\textit{Trump}" of person type and “\textit{Russia}" of geolocation type. These entities bring an interesting paradigm to the research, as now, for Fake News detection, we consider not only the news title content but also the association of multiple entities present in the news. 

To consider entities, our framework includes an entity encoder component, 
which first identifies the relevant entities in the news and then encodes them. We use Wikidata\footnote{https://www.wikidata.org/}, an open-source KG, as the source to match the entities and the ComplEx KG embedding technique \cite{trouillon2016complex} to embed the entities. Following our news example, “\textit{Trump}" from the news can be mapped to the respective KG entity instance, which in turn is connected to other entities in the form of triplets. A triplet is a collection of three elements that are represented in the \textit{(h, r, t)} fashion. Here,  \textbf{h} and \textbf{t} represent entities, and \textbf{r} represents the connection between them. An example of a connected triplet for the entity “\textit{Trump}" could be \textit{(Trump, birthCountry, USA)} and \textit{(Trump, gender, Male)}. We consider KG as the base of our entity encoder for the following reasons, (1) recent advances in KG embedding have shown efficient propagation of information within the graph, which makes an entity’s representation a consolidation of itself and also its neighbors' information, (2) several large scale KGs like Wikidata and DBpedia are open-source projects and hence readily available. The entity encoder component includes the following sub modules.

\subsubsection{Named entity recognition (NER)} this sub module 
assigns labels to the input news title. 
We use Spacy\footnote{https://spacy.io/} based RoBERTa model \cite{pritzkau2021nlytics}, which has shown comparative results with the state-of-the-art entity recognizer. The result is further passed to the disambiguation part.

\noindent \subsubsection{Named entity disambiguation (NED)} 
we include a disambiguation step, which connects the entity identified by NER with the most similar entity present in the KG. As we consider Wikidata KG as our knowledge base, we use disambiguation services\footnote{https://www.wikidata.org/w/api.php?action=help} exposed by Wikidata for this step. The result is the mapping of each entity from the last step with the corresponding Wikidata entity URI.

\subsubsection{KG embedding} we perform KG embedding using the ComplEx embedding algorithm \cite{trouillon2016complex}, which represents entities and relations of the KG in complex space. 
ComplEx embedding is favored as it can capture anti-symmetric relations better than the operations in the Euclidean space. 

\subsubsection{Entity encoder aggregation layer} the final sub module performs a permutation invariant aggregation of the entities' representation extracted after the KG embedding step. We use the mean operation to aggregate all the entities' embeddings present in a title. 

\subsection{Fake News classifier}

The Fake News classifier consolidated the representation output of the entity and news encoder sub modules. 
In our framework, the two representations are concatenated to create a super representation of the news title and entities. This representation is then passed to a feed forward network
where the final output represents the probability of the news as either true or fake. 
%
\section{Dataset Description and Experimental Setup}\label{sec:experiment}

In this section, we present the dataset, the baseline methods, and the experiment details. In its entirety, we want to answer the following questions,\\ 

\begin{itemize}

\item \textbf{Q1:} Is the DEAP-FAKED framework able to improve the Fake News detection performance by considering open-source KG?

\item \textbf{Q2:} What is the change in performance observed after considering entity information along with the news title?

\item \textbf{Q3:} How do the other open-source knowledge bases of textual nature, like Wikipedia, compare with KG for Fake News detection? 
\end{itemize}

\subsection{Dataset}

\subsubsection{Fake News dataset}
For a holistic analysis of our proposed framework, we considered news items belonging to diverse domains. The first dataset is the \textbf{Kaggle Fake News} dataset\footnote{https://www.kaggle.com/c/fake-news/overview}, which consists of 20,387 news items, having a near equal combination of True and Fake News. The news covers several domains, such as Politics, Business, and Technology. While the dataset provides several additional pieces of information, we ignore news content and author information and only consider the news title for our analysis. This decision further complicates the Fake News detection problem as the available resource for classification is quite limiting in terms of textual length. 
However, it is in accordance with previous studies observation that a majority of Fake News is propagated on social media platforms like Twitter, which has strict short text limits\footnote{https://news.mit.edu/2018/study-twitter-false-news-travels-faster-true-stories-0308}. 
Initial analysis of the dataset exposed the presence of bias terms in the dataset.
One example is the presence of a publication house name in the title of the news itself. Here most of the news items from famous publications like ``\textit{New York Times}" are true news.
These biases are usually introduced in the dataset during the data collection phase. To handle such cases, we removed any mention of the bias terms from the dataset. Other examples include the presence of certain politicians, celebrities, etc. 

Finally, only the news items whose entities can be mapped to the KG using NER and NED steps are kept. The complete pre-processing step brought down the news item count to $\sim$14k with a distribution of 60\% - 40\% of true and Fake News classes, respectively. We denote this dataset as \textbf{KFN-UB}. 

The second dataset is \textbf{CoAID} \cite{cui2020coaid}, which contains diverse COVID-19 healthcare misinformation, including Fake News from websites and social platforms. CoAID includes 4,251 news items. In accordance with the first dataset, we identified and removed bias terms from the CoAID dataset and then performed the text cleaning and entity mapping steps. This brought down the total news item count to 632. We denote this cleaned unbiased dataset as \textbf{CoAID-UB}. 

 

\subsubsection{Knowledge Graph}
For the KG, we use Wikidata5M \cite{wang2019kepler}, which is a subset of the Wikidata KG. It is created by only filtering the “valid” facts of Wikidata KG. The validity of a fact is confirmed if the entities and relations have a Wikipedia article with adequate description (at least five words long). 
As suggested in the previous section, the ComplEx algorithm is used to generate embedding for each entity in the Wikidata5M KG, which is later used in the entity encoder to filter out respective entities within the news.


\subsection{Baselines}
We compare DEAP-FAKED with the following models: 
\textbf{ExtraTreeClassifier:}  ExtraTreeClassifier is a decision tree-based classification algorithm that fits randomized decision trees on various sub-samples of the dataset. 
In our case, we first extract the count vectorization-based feature matrix from the tokenized news items and pass it to ExtraTreeClassifier.

\textbf{LSTM:} Long-Short Term Memory is a gated variant of Recurrent Neural Networks. 
In our case, we pass the title of the tokenized news items to the LSTM layer and connect the last hidden state of the LSTM with a sigmoid activated MLP layer to perform the Fake News classification.
   
\textbf{SentRoBERTa:} SentRoBERTa is a modification of the pre-trained RoBERTa network that uses siamese structure and triplet loss to derive semantically meaningful sentence embeddings. We use SentRoBERTa to generate sentence level embedding for news titles that are connected with the sigmoid activated MLP layer to perform the Fake News classification.

\textbf{StackedBiLSTM:} StackedBiLSTM is a two-layer stacking of the conventional bidirectional LSTM layer. 
In our case, we pass the tokenized news title ($t_1$, $t_2$ to $t_n$) to the StackedBiLSTM layer and connect the last hidden state with sigmoid activated MLP layers to perform the Fake News classification. 

\textbf{EntWiki-StackedBiLSTM:} This model incorporates the entities along with the news title. For each news item, the news title is encoded using the StackedBiLSTM, as discussed before. Apart from this, we leverage the entity encoder component of our framework with one major difference - instead of using KG, we use Wikipedia article material for entity encoding. For this purpose, we extract the Wikipedia description of the entity identified in the news item and encode that description using SentRoBERTa. The news title and entity encoding are then concatenated and passed to the sigmoid activated MLP layer to perform the Fake News classification.


\subsection{Experiment Setup}

\subsubsection{Metric} For evaluation of the Fake News detection, we consider Accuracy and F1 macro score as the preferred metrics of comparison. While accuracy is the de facto metric for the classification task, it lacks comparative prowess when the dataset is imbalanced. To handle such a case, we also consider the F1-score, which is a harmonic mean of the recall and precision metric. 

\subsubsection{Implementation details} Each model is developed and tested in Keras. For the performance calculation, each dataset has been split into an 80\% - 20\% ratio for train and test sets, respectively, and in a stratified fashion. For the KFN-UB dataset, the batch size is set to 32, whereas for the CoAID-UB dataset, it is 8. The hidden state’s dimension is fixed to 256 for all of the models, along with an early stop loss patience step size of 2 and max epochs of 100. For the bag of words model, the max feature size is set to 10k, and for the LSTM based models, the max vocabulary is set to 10k as well. Finally, each model is trained on the datasets for three trials with different seed values. The average performance metric is recorded for the best performing model on the test dataset.
\section{Evaluation}\label{sec:Eval}
\medskip
To answer the questions asked in Section \ref{sec:experiment}, we present the consolidated performance score in Table \ref{table:overallResult}.

\textbf{Q1} raises concerns about the use of DEAP-FAKED for improving the performance of Fake News detection. As evident by the results, DEAP-FAKED reports the highest score on both of the datasets (last row in Table \ref{table:overallResult}).
\begin{table}[!ht]
\begin{tabular}{l|c|c|c|c|llllll}
\cline{2-5}
 &
  \multicolumn{2}{c|}{\textbf{KFN-UB}} &
  \multicolumn{2}{c|}{\textbf{CoAID-UB}} &
  \multicolumn{1}{c}{\textbf{}} &
  \multicolumn{1}{c}{\textbf{}} &
  \multicolumn{1}{c}{\textbf{}} &
  \multicolumn{1}{c}{\textbf{}} &
  \multicolumn{1}{c}{\textbf{}} &
  \multicolumn{1}{c}{\textbf{}} \\ \cline{2-5}
Model vs Dataset \textbf{} &
  \textbf{F1 avg.} &
  \textbf{Acc avg.} &
  \textbf{F1 avg.} &
  \textbf{Acc avg.} &

   \\ \cline{1-5}
\multicolumn{1}{|l|}{\textit{ExtraTreeClassifier}}       & 0.77 & 0.78 & 0.75  & 0.77     \\ \cline{1-5}
\multicolumn{1}{|l|}{\textit{LSTM}}                      & 0.78 & 0.81 & 0.73  & 0.74     \\ \cline{1-5}
\multicolumn{1}{|l|}{\textit{SentRoBERTa}}               & 0.65 & 0.69  & 0.73 & 0.74      \\ \cline{1-5}
\multicolumn{1}{|l|}{\textit{StackedBiLSTM}}             & 0.79 & 0.81 & 0.75  & 0.76    \\ \cline{1-5}
\multicolumn{1}{|l|}{\textit{EntWiki-StackedBiLSTM}} & 0.88 & 0.89 & 0.74  & 0.75       \\ \cline{1-5}
\multicolumn{1}{|l|}{\textit{DEAP-FAKED}}         & \textbf{0.89} &  \textbf{0.90} & \textbf{0.78} & \textbf{0.78}  \\ \cline{1-5}
\end{tabular}
\caption{\textbf{Performance score of the models. For each of the datasets, we report F1 macro and Accuracy metric values. We present the average of the performance observed after performing 3 trials with different starting seeds. For both datasets, DEAP-FAKED reports the best performance value.}}
\label{table:overallResult}
\end{table}

\textbf{Q2} raises concerns over the requirement of entity information for the Fake News detection problem. To answer this question, we performed a comparative analysis of the models which contain the entity encoder module against the models which doesn’t. The result of this analysis is presented in Table \ref{table:q2result}. As evident from the table, models with the entity module have, on average higher performance scores for both datasets. To be exact, the average improvement is $\sim$13\% F1-score for KFN-UB and $\sim$1.5\% F1-score for CoAID-UB. 

\begin{table}[!htbp]
\centering
\begin{tabular}{l|l|l|l|l|}
\cline{2-5}
                                                         & \multicolumn{2}{c|}{\textbf{KFN-UB}} & \multicolumn{2}{c|}{\textbf{CoAID-UB}} \\ \cline{2-5} 
 & \multicolumn{1}{c|}{\textbf{F1 avg.}} & \multicolumn{1}{c|}{\textbf{Acc avg.}} & \multicolumn{1}{c|}{\textbf{F1 avg.}} & \multicolumn{1}{c|}{\textbf{Acc avg.}} \\ \hline
\multicolumn{1}{|l|}{\textit{Models w/o entity encoder}} & 0.7537            & 0.7805           & 0.7477             & 0.7585            \\ \hline
\multicolumn{1}{|l|}{\textit{Models w. entity encoder}}  & \textbf{0.8838}            & \textbf{0.8926}           & \textbf{0.7624}             & \textbf{0.7638}            \\ \hline
\end{tabular}
\caption{\textbf{Comparative score of models with and without the entity encoder sub module. We report the average F1 and Accuracy scores for KFN-UB and CoAID-UB datasets.}}
\label{table:q2result}
\end{table}

\textbf{Q3} raises concerns about the preference for a KG-based entity encoder in DEAP-FAKED against a text-based entity encoder. To answer this question, we can reflect back on the result of EntWiki-StackedBiLSTM and DEAP-FAKED from Table \ref{table:overallResult}. As evident in both cases, the KG-based model performs better than the text-based model. This is due to the concentrated information present in the KG-based embeddings, where encoding is propagated from multiple hops to the root entity node. This aggregation of relevant information lets the root node capture information which is far away in the KG. 
%
\section{Conclusions and Future Work}\label{sec:concl}
\medskip
Considering the recent surge in the number of Fake News, especially on online social media platforms, the topic related to Fake News detection has gained attention from the vast research community. In this work, we proposed DEAP-FAKED, a knowledge graph-based framework for the detection of Fake News. The approach uses minimum text, that is, only the title of the news articles, which requires less computational time and simulates the low-text Fake News propagation observed on the social media platform. We complement the low-text news title by identifying named entities and mapping these entities to an open-source KG. Embeddings are trained for entities in the KG by following an unsupervised KG embedding procedure, and the representation of the relevant entities is later filtered out for Fake News classification. On the dataset side, we consider a wide variety of datasets belonging to different domains. We carefully remove the bias from the datasets before feeding the data to the models. Comparing the proposed framework with other baseline approaches, we answer questions on the selection, performance, and preference of the proposed framework. Overall, DEAP-FAKED scores are better than the state-of-the-art results on both datasets.
Future work includes experimenting with a combination of KG-based and Wikipedia text base entity encoders that could lead to an enhanced framework.

\section*{Acknowledgment}
This research is financially
supported by EU H2020 SoBigData++ project (grant agreement No. 871042).

\bibliographystyle{IEEEtran}
\bibliography{main}

\end{document}